\def\year{2018}\relax
\documentclass[letterpaper]{article} 
\usepackage{aaai18}  
\usepackage{times}  
\usepackage{helvet}  
\usepackage{courier}  
\usepackage{url}  
\usepackage{graphicx}  
\usepackage{pgfplots}
\usetikzlibrary{backgrounds}
\pgfplotsset{compat=1.8}

\newcommand{\quotes}[1]{``#1''}

\begin{document}

%
\title{Teaching a Machine to Read Maps with Deep Reinforcement Learning}
 \author{Gino Brunner \and Oliver Richter \and Yuyi Wang \and Roger Wattenhofer\thanks{Names in alphabetic order}\\
ETH Zurich\\
\{brunnegi, richtero, yuwang, wattenhofer\}@ethz.ch
 }

\maketitle

\begin{abstract}
The ability to use a 2D map to navigate a complex 3D environment is quite remarkable, and even difficult for many humans. Localization and navigation is also an important problem in domains such as robotics, and has recently become a focus of the deep reinforcement learning community. In this paper we teach a reinforcement learning agent to read a map in order to find the shortest way out of a random maze it has never seen before. Our system combines several state-of-the-art methods such as A3C and incorporates novel elements such as a recurrent localization cell. Our agent learns to localize itself based on 3D first person images and an approximate orientation angle. The agent generalizes well to bigger mazes, showing that it learned useful localization and navigation capabilities. 
\end{abstract}

\section{Introduction}

One of the main success factors of human evolution is our ability to craft and use complex
tools. Not only did this ability give us a motivation for social interaction by teaching others how to use different tools, it also enhanced
our thinking capabilities, since we had to understand ever more complex tools. Take a map as an example; a map helps us navigate places we have never seen before. However, we first need to learn how to read it, i.e., we need to associate the content of a two-dimensional map with our three-dimensional surroundings. 
With algorithms becoming increasingly capable of learning complex relations, a way to make machines intelligent is to teach them how to use already existing
tools. In this paper, we teach a machine how to read a map with deep reinforcement learning.

The agent wakes up in a maze. The agent's view is an image: the maze rendered from the agent's perspective, like a dungeon in a first person video game. This rendered image is provided by the DeepMind Lab environment \cite{DBLP:journals/corr/BeattieLTWWKLGV16}.
The agent can be controlled by a human, or as in our case, by a complex deep reinforcement learning architecture.\footnote{Our code can be found here: \url{https://github.com/OliverRichter/map-reader.git}}
The agent can move (forward, backward, left, right) and rotate (left, right), and its view image will change accordingly. In addition, the agent gets to see a map of the maze, also an image, as can be seen in Figure~\ref{fig:overview}.
One location on the map is marked with an ``X'' –- the agent's target. The crux is that the agent does not know where on the map it currently is. Several locations on the map might correspond well with the current view. Thus the agent needs to move around to learn its position and then move to the target, as illustrated in Figures \ref{fig:trajectories} and \ref{fig:localization}. 
We do equip the agent with an approximate orientation angle, i.e., the agent roughly knows the direction it is moving or looking. In the map, up is always north. During training the agent learns which approximate orientation corresponds to north.

A complex multi-stage task, such as navigating a maze with the help of a map, can be naturally decomposed into several subtasks: (i) The agent needs to observe its 3D environment and compare it to the map to determine its most likely position.
(ii) The agent needs to understand the map, or in our case associate symbols on the map with rewards and thereby gain an understanding of what a wall is, what navigable space is, and what the target is. 
(iii) Finally the agents needs to learn how to follow a plan in order to reach the target.

Our contribution is as follows: We present a novel modular reinforcement learning architecture that consists of a reactive agent and several intermediate subtask modules. Each of these modules is designed to solve a specific subtask. 
The modules themselves can contain neural networks or alternatively implement exact algorithms or heuristics. Our presented agent is capable of finding the target in random mazes roughly three times the size of the largest mazes it has seen during training.

Further contributions include:
\begin{itemize}
\item The Recurrent Localization Cell that outputs a location probability distribution based on an estimated stream of visible local maps. 
\item A simple mapping module that creates a visible local 2D map from 3D RGB input. The mapping module is robust, even if the agent's compass is inaccurate.
\end{itemize}

\section{Related Work}

Reinforcement learning in relation to AI has been studied since the 1950's \cite{minsky1954theory}. 
Important early work on reinforcement learning includes the temporal difference learning method by \citeauthor{sutton1984temporal} \shortcite{sutton1984temporal,DBLP:journals/ml/Sutton88}, which is the basis for actor-critic algorithms \cite{DBLP:journals/tsmc/BartoSA83} and Q-learning techniques \cite{watkins1989learning,watkins1992q}.
First works using artificial neural networks for reinforcement learning include \cite{DBLP:journals/ml/Williams92} and \cite{DBLP:journals/nn/Gullapalli90}.
For an in-depth overview of reinforcement learning we refer the interested readers to \cite{DBLP:journals/jair/KaelblingLM96},  \cite{DBLP:books/lib/SuttonB98}  
and \cite{DBLP:series/synthesis/2010Szepesvari}.

The current deep learning boom was started by, among other contributions, the \emph{backpropagation} algorithm \cite{rumelhart1988learning} and advances in computing power and GPU frameworks. 
However, deep learning could not be applied effectively to reinforcement learning until recently. \citeauthor{DBLP:journals/nature/MnihKSRVBGRFOPB15}~\shortcite{DBLP:journals/nature/MnihKSRVBGRFOPB15} introduced the Deep-Q-Network (DQN) that uses experience replay and target networks to stabilize the learning process. 
Since then, several extensions to the DQN architecture have been proposed, such as the Double Deep-Q-Network (DDQN) \cite{DBLP:conf/aaai/HasseltGS16} and the dueling network architecture \cite{DBLP:conf/icml/WangSHHLF16}. 
These networks are based on using replay buffers to stabilize learning, such as prioritized experience replay \cite{DBLP:journals/corr/SchaulQAS15}. 
The state-of-the-art A3C \cite{DBLP:journals/corr/MnihBMGLHSK16} relies on asynchronous actor-learners to stabilize learning. 
In our system, we use A3C learning on a modified network architecture to train our reactive agent and the localization module in an on-policy manner. We also make use of (prioritized) replay buffers to train our agent off policy.

A major challenge in reinforcement learning are environments with delayed or sparse rewards. An agent that never gets a reward can never learn good behavior. 
Thus \citeauthor{DBLP:journals/corr/JaderbergMCSLSK16} \shortcite{DBLP:journals/corr/JaderbergMCSLSK16} and \citeauthor{DBLP:journals/corr/MirowskiPVSBBDG16} \shortcite{DBLP:journals/corr/MirowskiPVSBBDG16}
introduced auxiliary tasks that let the agent learn based on intermediate intrinsic pseudo-rewards,
such as predicting the depth from a 3D RGB image, while simultaneously trying to solve the main task, e.g., finding the exit in a 3D maze. The policies learned by the auxiliary tasks are not directly used by the agent, but solely serve the purpose of helping the agent learn better representations which improves its performance on the main task. The idea of auxiliary tasks is inspired by prior work on temporal abstractions, such as options \cite{DBLP:journals/ai/SuttonPS99}, whose focus was on learning temporal abstractions to improve high-level learning and planning. In our work we introduce a modularized architecture that incorporates intermediate subtasks, such as localization, local map estimation and global map interpretation. In contrast to \cite{DBLP:journals/corr/JaderbergMCSLSK16}, our reactive agent directly uses the outputs of these modules to solve the main task. Note that we use an auxiliary task inside our localization module to improve the local map estimation. 
\citeauthor{DBLP:conf/nips/KulkarniNST16} \shortcite{DBLP:conf/nips/KulkarniNST16} introduced a hierarchical version of the DQN to tackle the challenge of delayed and sparse rewards. Their system operates at different temporal scales and allows the definition of goals using entity relations. The policy is learned in such a way to reach these goals. We use a similar approach to make our agent follow a plan, such as, \quotes{go north}. 

Mapping and localization has been extensively studied in the domain of robotics \cite{thrun2005probabilistic}. A robot creates a map of the environment from sensory input (e.g., sonar or LIDAR) and then uses this map to plan a path through the environment. 
Subsequent works have combined these approaches with computer vision techniques \cite{DBLP:journals/air/Fuentes-PachecoAR15} that use RGB(-D) images as input. 
Machine learning techniques have been used to solve mapping and planning separately, and later also tackled the joint mapping and planning problem \cite{DBLP:journals/computer/Elfes89}. Instead of separating mapping and planning phases, reinforcement learning methods aimed at directly learning good policies for robotic tasks, e.g., for learning human-like motor skills \cite{DBLP:journals/nn/PetersS08}. 

Recent advances in deep reinforcement learning have spawned impressive work in the area of mapping and localization. 
The UNREAL agent \cite{DBLP:journals/corr/JaderbergMCSLSK16} uses auxiliary tasks and a replay buffer to learn how to navigate a 3D maze. 
\citeauthor{DBLP:journals/corr/MirowskiPVSBBDG16} \shortcite{DBLP:journals/corr/MirowskiPVSBBDG16} came up with an agent that uses different auxiliary tasks in an online manner to understand if navigation capabilities manifest as a bi-product of solving a reinforcement learning problem. 
\citeauthor{DBLP:conf/icra/ZhuMKLGFF17} \shortcite{DBLP:conf/icra/ZhuMKLGFF17} tackled the problems of generalization across tasks and data inefficiency. 
They use a realistic 3D environment with physics engine to gather training data efficiently. Their model is capable of navigating to a visually specified target. 
In contrast to other approaches, they use a memory-less feed-forward model instead of recurrent models. 
\citeauthor{DBLP:journals/corr/GuptaDLSM17} \shortcite{DBLP:journals/corr/GuptaDLSM17} simulated a robot that navigates through a real 3D environment. 
They focus on the architectural problem of learning mapping and planning in a joint manner, such that the two phases can profit from knowing each other's needs. 
Their agent is capable of creating an internal 2D representation of the local 3D environment, similar to our local visible map. 
In our work a global map is given, and the agent learns to interpret and read that map to reach a certain target location. 
Thus, our agent is capable of following complicated long range trajectories in an approximately shortest path manner. 
Furthermore, their system is trained in a fully supervised manner, whereas our agent is trained with reinforcement learning. 
\citeauthor{DBLP:journals/corr/BhattiDMNST16}~\shortcite{DBLP:journals/corr/BhattiDMNST16}  augment the standard DQN with semantic maps in the VizDoom~\cite{DBLP:conf/cig/KempkaWRTJ16} environment. These semantic maps are constructed from 3D RGB-D input, and they employ techniques such as standard computer vision based object recognition and SLAM. 
They showed that this results in better learned policies. 
The task of their agent is to eliminate as many opponents as possible before dying. 
In contrast, our agent needs to escape from a complex maze. 
Furthermore, our environments are designed to provide as little semantic information as possible to make the task more difficult for the agent; our agent needs to construct its local visible map based purely on the shape of its surroundings.

\section{Architecture}

Many complex tasks can be divided into easier intermediate tasks which when all solved individually solve the complex task. We use this principle and apply it to neural network architecture design. 
In this section we first introduce our concept of modular intermediate tasks, and then discuss how we implement the modular tasks in our map reading architecture.

\subsection{Modular Intermediate Tasks}

An \textit{intermediate task module} can be any information processing unit that takes as input either sensory input and/or the output of other modules. A module is defined and designed after the intermediate task it solves and can consist of trainable and hard coded parts. Since we are dealing with neural networks, the output and therefore the input of a module can be erroneous. Each module adjusts its trainable parameters to reduce its error independent of other modules. We achieve this by stopping error back-propagation on module boundaries. Note that this separation has some advantages and drawbacks:
\begin{itemize}
\item Each module performance can be evaluated and debugged individually.
\item Small intermediate subtask modules have short credit assignment paths, which reduces the problem of exploding and vanishing gradients during back-propagation.
\item Modules cannot adjust their output to fit the input needs of the next module. This has to be achieved through interface design, i.e., intermediate task specification.
\end{itemize}

\begin{figure}[t]
\centering
\includegraphics[width=0.9\columnwidth]{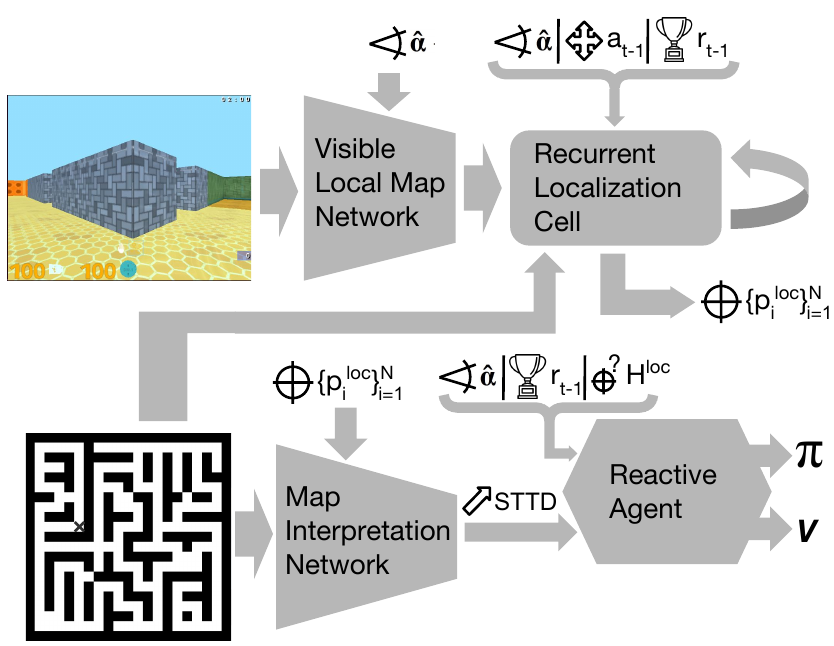}
\caption{Architecture overview and interplay between the four modules. $\hat{\alpha}$ is the discretized angle, $a_{t-1}$ is the last action taken, $r_{t-1}$ is the last reward received, $\{p_i^{loc}\}_{i=1}^N$ is the estimated location probability distribution over the $N$ possible discrete locations, $H^{loc}$ is the entropy of the estimated location probability distribution, STTD is the short term target direction suggested by the map interpretation network, $V$ is the estimated state value and $\pi$ is the policy output from which the next action $a_t$ is sampled.}
\label{fig:overview}
\end{figure}

Our neural network architecture consists of four modules, each dedicated to a specific subtask. We first give an overview of the interplay between the modules before describing them in detail in the following sections. The architecture overview is sketched in Figure~\ref{fig:overview}.

The first module is the \textit{visible local map network}; it takes the raw visual input from the 3D environment and creates for each frame a two dimensional map excerpt of the currently visible surroundings. The second module, the \textit{recurrent localization cell}, takes the stream of visible local map excerpts and integrates it into a local map estimation. This local map estimation is compared to the global map to get a probability distribution over the discretized possible locations. The third module is called \textit{map interpretation network}; it learns to interpret the global map and outputs a short term target direction for the estimated position. The last module is a \textit{reactive agent} that learns to follow the estimated short term target direction to ultimately find the exit of the maze.

We allow our agent to have access to a discretized angle $\hat{\alpha}$ describing the direction it is facing, comparable to a robot having access to a compass. Furthermore, we do not limit ourself to completely unsupervised learning and allow the agent to use a discretized version of its actual position during training. This could be implemented as a robot training on the network with the help of a GPS signal. The robot could train as long as the accuracy of the GPS signal is below a certain threshold and act on the trained network as soon as the GPS signal gets inaccurate or totally lost. We leave such a practical implementation of our algorithm to future work and focus here on the algorithmic structure itself.

We now describe each module architecture individually before we discuss their joint training in Section~\ref{training}. If not specified otherwise, we use rectified linear unit activations after each layer.

\subsection{Visible Local Map Network}
\label{vlm}

\begin{figure}[t]
\includegraphics[width=0.95\columnwidth]{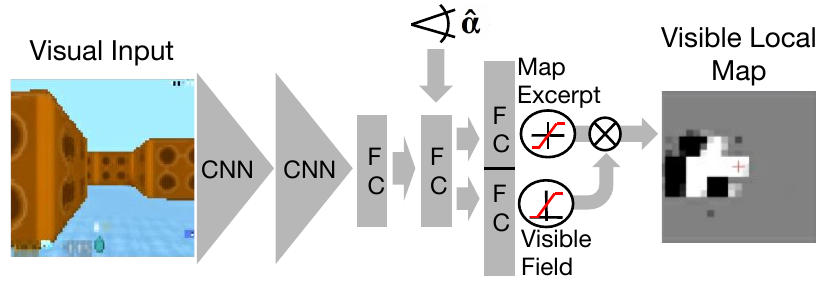}
\caption{The visible local map network: The RGB pixel input is passed through two convolutional neural network (CNN) layers and a fully connected (FC) layer before being concatenated to the discretized angle $\hat{\alpha}$ and further processed by fully connected layers and a gating operation.}
\label{fig:vlm}
\end{figure}

The visible local map network preprocesses the raw visual RGB input from the environment through two convolutional neural network layers followed by a fully connected layer. We adapted this preprocessing architecture from \cite{DBLP:journals/corr/JaderbergMCSLSK16}. The thereby generated features are concatenated to a 3-hot discretized encoding $\hat{\alpha}$ of the orientation angle $\alpha$, i.e., we input the angle as $n$-dimensional vector where each dimension represents a discrete state of the angle, with $n=30$. We set the three vector components that represent the discrete angle values closest to the actual angle to one while the remaining components are set to zero, e.g. $\hat{\alpha} = [0 \ldots 0 1 1 1 0 \ldots 0]$. We used a 3-hot instead of a 1-hot encoding to smooth the input.
Note that this encoding has an average quantization error of 6 degrees.

The discretized angle and preprocessed visual features are passed through a fully connected layer to get an intermediate representation from which two things are estimated:
\begin{enumerate}
\item A reconstruction of the map excerpt that corresponds to the current visual input
\item The current field of view, which is used to gate the estimated map excerpt such that only estimates which lie in the line of sight make it into the visible local map. This gating is crucial to reduce noise in the visible local map output. 
\end{enumerate}

See Figure~\ref{fig:vlm} for a sketch of the visible local map network architecture.

\begin{figure}[t]
\centering
\includegraphics[width=0.85\columnwidth]{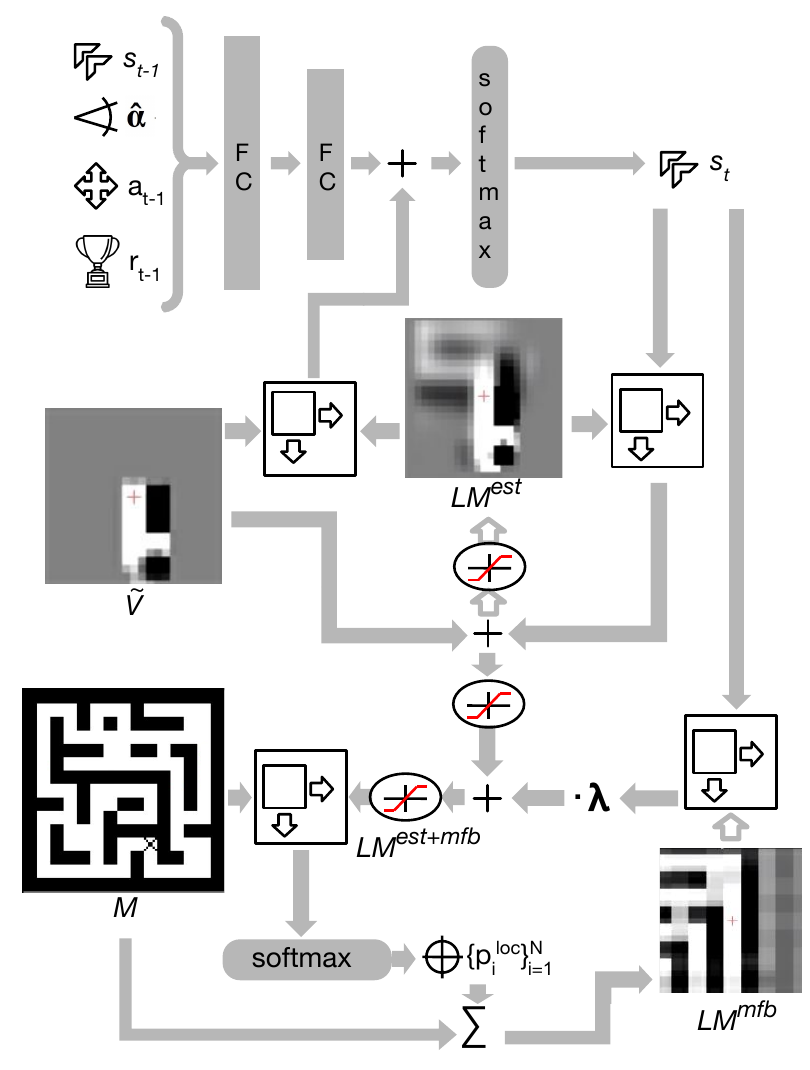}
\caption{Sketch of the information flow in the recurrent localization cell. The last egomotion estimation $s_{t-1}$, the discretized angle $\hat{\alpha}$, the last action $a_{t-1}$ and reward $r_{t-1}$ are passed through two fully connected (FC) layers and combined with a two dimensional convolution between the former local map estimation $LM_{t-1}^{est}$ and the current visible local map input to get the new egomotion estimation $s_t$. This egomotion estimation is used to shift the previously estimated local map $LM_{t-1}^{est}$ and the previous map feedback local map $LM_{t-1}^{mfb}$. A weighted and clipped combination of these local map estimations, $LM_{t-1}^{est+mfb}$, is convolved with the full map to get the estimated location probability distribution $\{p_i^{loc}\}_{i=1}^N$. 
Recurrent connections are marked by empty arrows.
}
\label{fig:rlc}
\end{figure}

\subsection{Recurrent Localization Cell}
\label{rlc}

Moving around in the environment, the agent generates a stream of visible local map excerpts like the output in Figure~\ref{fig:vlm} or the visible local map input $\tilde{V}$ in Figure~\ref{fig:rlc}. The recurrent localization cell then builds an egocentric local map out of this stream and compares it to the actual map to estimate the current position. The agent has to predict its egomotion to shift the egocentric estimated local map accordingly. We refer to Figure~\ref{fig:rlc} for a sketch of the architecture described hereafter.

Let $M$ be the current map, $\tilde{V}$ the output of the visible local map network, $\hat{\alpha}$ the discretized 3-hot encoded orientation angle, $a_{t-1}$ the 1-hot encoded last action taken, $r_{t-1}$ the extrinsic reward received by taking action $a_{t-1}$, $LM_t^{est}$ the estimated local map at time step $t$, $LM_t^{mfb}$ the map feedback local map at time step $t$, $LM_t^{est+mfb}$ the estimated local map with map feedback at time step $t$, $s_t$ the estimated necessary shifting (or estimated egomotion) at time step t and $\{p_i^{loc}\}_{i=1}^N$ the discrete estimated location probability distribution. Then we can describe the functionality of the recurrent localization cell by the following equations:

\[s_t = softmax(f(s_{t-1},\hat{\alpha},a_{t-1},r_{t-1}) + LM_{t-1}^{est} * \tilde{V}) \]
\[LM_t^{est}=\left[LM_{t-1}^{est}*s_t+\tilde{V}\right]_{-0.5}^{+0.5}\]
\[LM_t^{est+mfb} = \left[LM_t^{est} + \lambda\cdot LM_{t-1}^{mfb}*s_t\right]_{-0.5}^{+0.5}\]
\[\{p_i^{loc}\}_{i=1}^N = softmax\left(m*LM_t^{est+mfb}\right)\]
\[LM_t^{mfb} = \sum_{i=1}^N p_i\cdot g(m,i)\]
Here, $f(\cdot)$ is a two layer feed forward neural network, $*$ denotes a two dimensional discrete convolution with stride one in both dimensions, $[\cdot]_{-0.5}^{+0.5}$ denotes a clipping to $[-0.5,+0.5]$, $\lambda$ is a trainable map feedback parameter
and $g(m,i)$ extracts from the map $m$ the local map around location $i$. 

\subsection{Map Interpretation Network}

The goal of the map interpretation network is to find rewarding locations on the map and construct a plan to get to these locations. We achieve this in three stages: First, the network passes the map through two convolutional layers followed by a rectified linear unit activation to create a 3-channel reward map. The channels are trained (as discussed in Section~\ref{training}) to represent wall locations, navigable locations and target locations respectively. This reward map is then area averaged, rectified and passed to a parameter free 2D shortest path planning module which outputs for each of the discrete locations on the map a distribution over \{North, East, South, West\}, i.e., a \textit{short term target direction} (STTD), as well as a measure of distance to the nearest target location. This plan is then multiplied with the estimated location probability distribution to get the smooth STTD and target distance of the currently estimated location. Note that planning for each possible location and querying the plan with the full location probability distribution helps to resolve the exploitation-exploration dilemma of the reactive agent:
\begin{itemize}
\item An uncertain location probability distribution close to the uniform distribution will result in an uncertain STTD distribution over \{North, East, South, West\}, thereby encouraging exploration.
\item A location probability distribution over locations with similar STTD will accumulate these similarities and result in a clear STTD for the agent, even though the location might still be unclear (exploitation).
\end{itemize}

\subsection{Reactive Agent and Intrinsic Reward}

As mentioned, the reactive agent faces two partially contradicting goals: following the STTD (exploitation) and improving the localization by generating information rich visual input (exploration), e.g., no excessive staring at walls. The agent learns this trade off through reinforcement learning, i.e., by maximizing the expected sum of rewards. The rewards we provide here are extrinsic rewards from the environment (negative reward for running into walls, positive reward for finding the target) as well as intrinsic rewards linked to the short term goal inputs of the reactive agent. These short term goal inputs are the STTD distribution over \{North, East, South, West\} and the measure of distance to the nearest target location from the map interpretation network as well as the normalized entropy $H^{loc}$ of the discrete location probability distribution $\{p_i^{loc}\}_{i=1}^N$.
$H^{loc}$ represents a measure of location uncertainty which is linked to the need for exploration.

The intrinsic reward consists of two parts to encourage both exploration and exploitation. The exploration intrinsic reward $I_t^{explor}$ in each timestep $t$ is the difference in location probability distribution entropy to the previous timestep: 

\[I_t^{explor} = H_{t-1}^{loc} - H_t^{loc}\]

Note that this reward is positive if and only if the location probability distribution entropy decreases, i.e., when the agent gets more certain about its position.

The exploitation intrinsic reward should be a measure of how well the egomotion of the agent aligns with the STTD. For this we calculate an approximate two dimensional egomotion vector $\vec{e}_t$ from the egomotion probability distribution estimation $s_t$.
Similarly we calculate a STTD vector $\vec{d}_{t-1}$ from the STTD distribution over $\{North, East, South, West\}$ of the previous timestep. We calculate the exploitation intrinsic reward $I_t^{exploit}$ as dot product between the two vectors: 

\[I_t^{exploit} = \vec{e_t}^T\cdot\vec{d}_{t-1}\]

Note that this reward is positive if and only if the angle difference between the two vectors is no bigger than 90 degrees, i.e., if the estimated egomotion was in the same direction as suggested by the STTD in the timestep before.

As input to the reactive agent we concatenate the discretized 3-hot angle $\hat{\alpha}$, the last extrinsic reward and the location probability distribution entropy $H^{loc}$ to the STTD distribution and the estimated target distance. The agent itself is a simple feed-forward network consisting of two fully connected layers with rectified linear unit activation followed by a fully connected layer for the policy and a fully connected layer for the estimated state value respectively. The agents next action is sampled from the softmax-distribution over the policy outputs.

\subsection{Training Losses}
\label{training}

To train our agent, we use a combination of on-policy losses, where the data is generated from rollouts in the environment, and off-policy losses, where we sample the data from a replay memory. More specifically, the total loss is the sum of the four module specific losses:
\begin{enumerate}
\item $L^{vlm}$, the off-policy visible local map loss 
\item $L^{loc}$, the on-policy localization loss
\item $L^{rm}$, the off-policy reward map loss and
\item $L^{a}$, the on-policy reactive agents acting loss
\end{enumerate}
We train our agent as asynchronous advantage actor critic, or A3C, with additional losses; similar to DeepMind's UNREAL agent \cite{DBLP:journals/corr/JaderbergMCSLSK16}:

In each training iteration, every thread rolls out up to 20 steps in the environment and accumulates the localization loss $L^{loc}$ and acting loss $L^{a}$. For each step, an experience frame is pushed to an experience history buffer of fixed length. Each experience frame contains all inputs the network requires as well as the current discretized true position.
From this experience history, frames are sampled and inputs replayed through the network to calculate the visible local map loss $L^{vlm}$ and the reward map loss $L^{rm}$. We now describe each loss in more detail.

The output $\tilde{V}$ of the visible local map network is trained to match the visible excerpt of the map $V$, constructed from the discretized location and angle. 
In each training iteration 20 experience frames are uniformly sampled from the experience history and the visible local map loss is calculated as the sum of L2 distances between visible local map outputs $\tilde{V}_k$ and targets $V_k$:

\[L^{vlm} = \sum_{k\in\mathcal{S}}||\tilde{V}_k - V_k||_2\]

Here, $\mathcal{S}$ denotes the set of sampled frame indices.

Our localization loss $L_{loc}$ is trained on the policy rollouts in the environment. For each step, we compare the estimated position to the actual position in two ways, which results in a cross entropy location loss $L^{loc,xent}$ and a distance location loss $L^{loc,d}$. The cross entropy location loss is the cross entropy between the location probability distribution $\{p_i^{loc}\}_{i=1}^N$ and a 1-hot encoding of the actual position. The distance loss $L^{loc,d}$ is calculated at each step as the L2 distance between the actual two dimensional cell position coordinates $\vec{c}_{pos}$ and the estimated centroid of all possible cells $i$ weighted by their corresponding probability $p_i^{loc}$:

\[L^{loc,d} = \left|\left|\vec{c}_{pos} - \sum_{i=1}^N p_i^{loc} \cdot\vec{c}_i\right|\right|_2\]

In addition to training the location estimation directly we also assign an auxiliary local map loss $L^{loc,lm}$ to help with the local map construction. We calculate the local map loss only once per training iteration as L2 distance between the last estimated local map $LM^{est}$ and the actual local map at that point in time.

\begin{figure}[t]
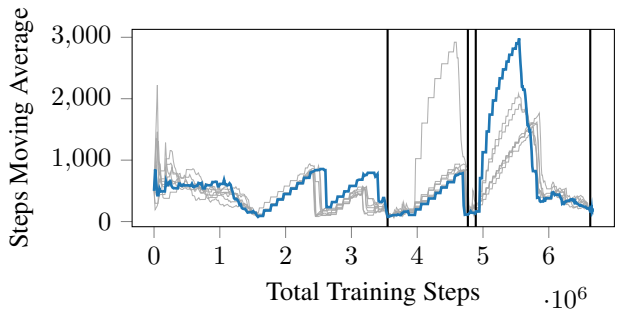

\centering
\include{average_steps_during_training}
\caption{Training performance of 8 actor threads that start training on 5x5 mazes. The vertical black lines mark jumps to larger mazes of the thread in blue.}
\label{fig:avgStepsTraining}
\end{figure}

The goal of the reward map loss $L^{rm}$ is to have the three channels of the reward map represent wall locations, free space locations and target locations respectively. To do this, we leverage the setting that running into a wall gives a negative extrinsic reward, moving in open space gives no extrinsic reward and finding the target gives a positive extrinsic reward. Therefore the problem can be transformed into estimating an extrinsic reward. Each training iteration we sample 20 frames from the experience history. This sampling is independent from the visible local map loss sampling and skewed to have in expectation equally many frames with positive, negative and zero extrinsic reward. 
For each frame, the frames map is passed through the convolution layers of the map interpretation network to create the corresponding reward map while the visual input and localization state saved in the frame are fed through the network to get the estimated location probability distribution. The reward map loss is the cross entropy prediction error of the reward at the estimated position.

Our reactive agent's acting loss is equivalent to the A3C learning described by \citeauthor{DBLP:journals/corr/MnihBMGLHSK16}~\shortcite{DBLP:journals/corr/MnihBMGLHSK16}. We also adapted an action repeat of 4 and a frame rate of 15 fps.
The whole network is trained by RMSprop gradient descent with gradient back propagation stopped at module boundaries, i.e., each module is only trained on its module specific loss.

\section{Environment and Results}

\begin{table*}[ht]
\centering
\begin{tabular}{|l|c|c|c|c|c|c|c|c|c|}
\hline
Maze size & 5x5&7x7&9x9&11x11&13x13&15x15&17x17&19x19&21x21\\ \hline
Targets found &100\%&100\%&100\%&99\%&99\%&98\%&93\%&93\%&91\%\\
\hline
\end{tabular}%
\caption{Percentage of targets found in the test mazes. Up to size 9x9 the agent always finds the target. More interestingly, the agent is able to find more than 90\% of the targets in mazes that are bigger than any maze it has seen during training.}
\label{tab:percentage_found}
\end{table*}

To evaluate our architecture we created a training and test set of mazes with the corresponding black and white maps in the DeepMind Lab environment. The mazes are quadratic grid mazes with each maze cell being either a wall, an open space, the target or the spawn position. The training set consists of 100 mazes of different sizes; 20 mazes each in the sizes 5x5, 7x7, 9x9, 11x11 and 13x13 maze cells. The test set consists of 900 mazes; 100 in each of the sizes 5x5, 7x7, 9x9, 11x11, 13x13, 15x15, 17x17, 19x19 and 21x21.
Note that the outermost cells in the mazes are always walls, therefore the maximal navigable space of a 5x5 maze is 3x3 maze cells. Thus the navigable space for the biggest test mazes is roughly 3 times larger than for the biggest training mazes.

For the localization, we used a location cell granularity 3 times finer than the maze cells, which results in a total of $N$=63x63=3969 discrete location states on the biggest 21x21 mazes. 
We train our agent starting on small mazes and increase the maze sizes as the agent gets better. More specifically we use 16 asynchronous agent training threads from which we start 8 on the smallest (5x5) training mazes while the other training threads are started 2 each on the other sizes (7x7, 9x9, 11x11 and 13x13). This prevents the visible local map network from overfitting on the small 5x5 mazes. The thread agents are placed into a randomly sampled maze of their currently associated maze size and try to find the exit, while counting their steps. 
A \emph{step} is one interaction with the environment, i.e., sampling an action from the agents policy $\pi$ and receiving the corresponding next visual input, discretized angle and extrinsic reward from the environment. A step is not the same as a location or maze grid cell; as agents accelerate, there is no direct correlation between steps and actual walked distance. We consider each sampled maze an episode start. The episode ends successfully if the agent manages to find the target and the steps needed are stored. If the agent does not find the exit in 4500 steps, the episode ends as not successful. After an episode ends, a new episode is started, i.e., a new maze is sampled. Note that in this setting the agent is always placed in a newly sampled maze and not in the same maze as in \cite{DBLP:journals/corr/JaderbergMCSLSK16} and \cite{DBLP:journals/corr/MirowskiPVSBBDG16}.

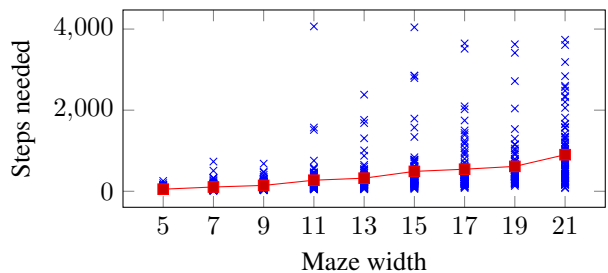
\begin{figure}[t]
\centering
\begin{tikzpicture}
	\begin{axis}[ylabel={Steps needed},xlabel={Maze width},xtick={5,7,9,11,13,15,17,19,21}, width = 0.95\columnwidth,
height = 0.5\columnwidth,]
		\addplot plot [only marks,mark=x] file {results/steps.csv};
        \addplot plot [] file {results/averages.csv};
	\end{axis}
\end{tikzpicture}
\caption{All the results of the (at most 100) successful tests for each maze size. Every single test is represented by an ``x''. The line connects the arithmetic averages of each maze size. The distance between origin and target grows linearly with maze size, as does the number of steps.}
\label{fig:full_agent_test_performance}
\end{figure}

For each thread we calculate a moving average of steps needed to end the episodes. Once this moving average falls below a maze size specific threshold, the thread is transferred to train on mazes of the next bigger size. Once a thread's moving average of steps needed in the biggest training mazes (13x13) falls below the threshold, the thread is stopped and its training is considered successful. Once all threads reach this stage, the overall training is considered successful and the agent is fully trained. We calculate the moving average over the last 50 episodes and use 60, 100, 140, 180 and 220 steps as threshold for the maze sizes 5x5, 7x7, 9x9, 11x11 and 13x13, respectively. Figure \ref{fig:avgStepsTraining} shows the training performance of 8 actor threads. One can see that the agents sometimes overfit their policies which results in temporarily decreased performance even though the maze size did not increase. In the end however, all threads reach good performance. 

\begin{figure}[t]
\centering
\includegraphics[width=0.95\columnwidth]{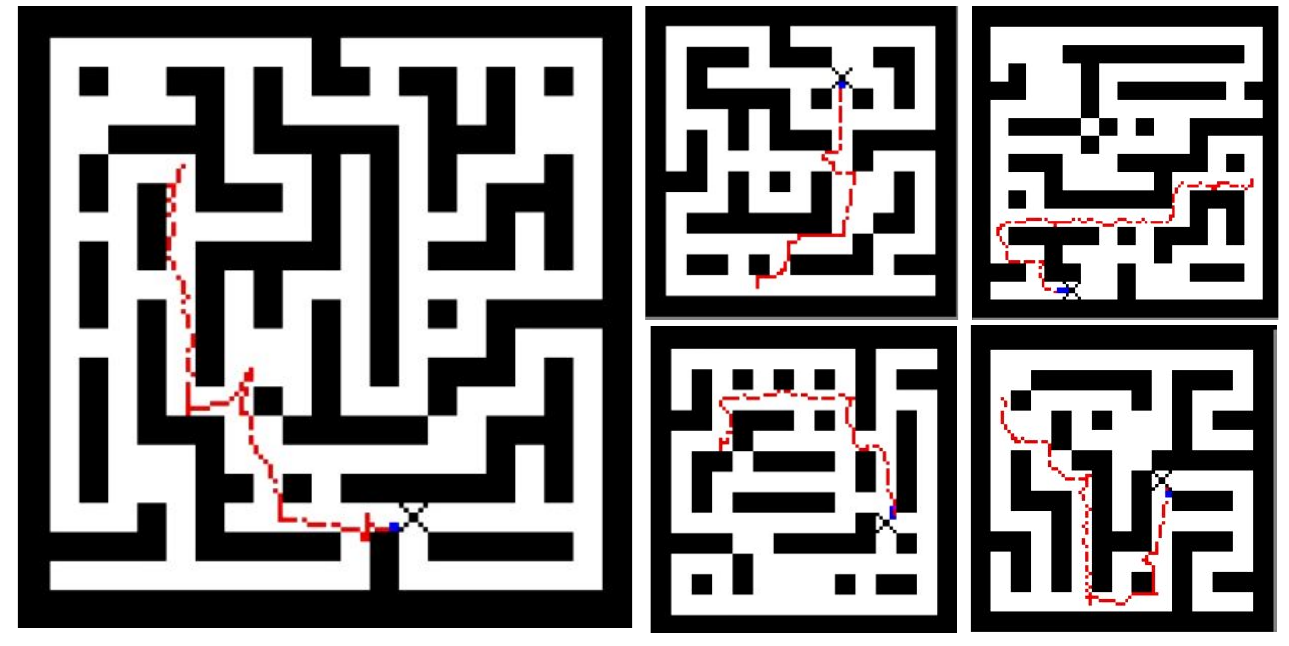}
\caption{Example trajectories walked by the agent. Note that the agent walks close to the shortest path and its continuous localization and planning lets the agent find the path to the target even after it took a wrong turn.}
\label{fig:trajectories}
\end{figure}

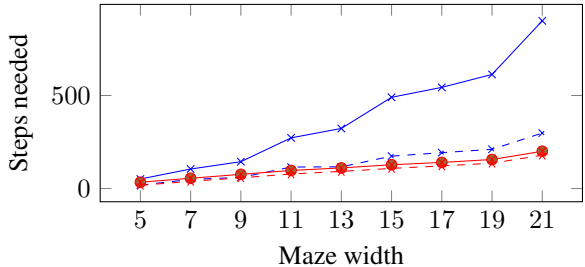
\begin{figure}[t]
\centering
\begin{tikzpicture}
	\begin{axis}[ylabel={Steps needed},xlabel={Maze width},xtick={5,7,9,11,13,15,17,19,21}, width = 0.95\columnwidth,
height = 0.5\columnwidth,]
        \addplot plot [blue, mark=x] file {results/averages.csv};
        \addplot plot [blue,dashed, mark=x] file {results/averages_without_turning.csv};
        \addplot plot [red] file {results/opt_averages.csv};
        \addplot plot [red,dashed] file {results/opt_averages_without_turning.csv};
	\end{axis}
\end{tikzpicture}
\caption{Comparison of our agent (blue lines) to an  agent that has perfect position information and an optimal short term target direction input (red lines). The solid lines count all steps (turns and moves). The solid blue line is the same as the average line of Figure \ref{fig:full_agent_test_performance}. The dashed lines do not count the steps in which the agent turns. The figure shows that the overhead is mostly because of turning, as our agent needs to ``look around'' to localize itself.}
\label{fig:full_agent_test_comparison}
\end{figure}

The trained agent is tested on the 900 test set mazes, the number of required steps per maze size are plotted in Figure~\ref{fig:full_agent_test_performance}. We stop a test after 4,500 steps, but even for the biggest test mazes (21x21) the agent found more than 90\% of the targets within these 4,500 steps. See Table~\ref{tab:percentage_found} for the percentage of exits found in all maze sizes.

\begin{figure}[t]
\centering
\includegraphics[width=0.95\columnwidth]{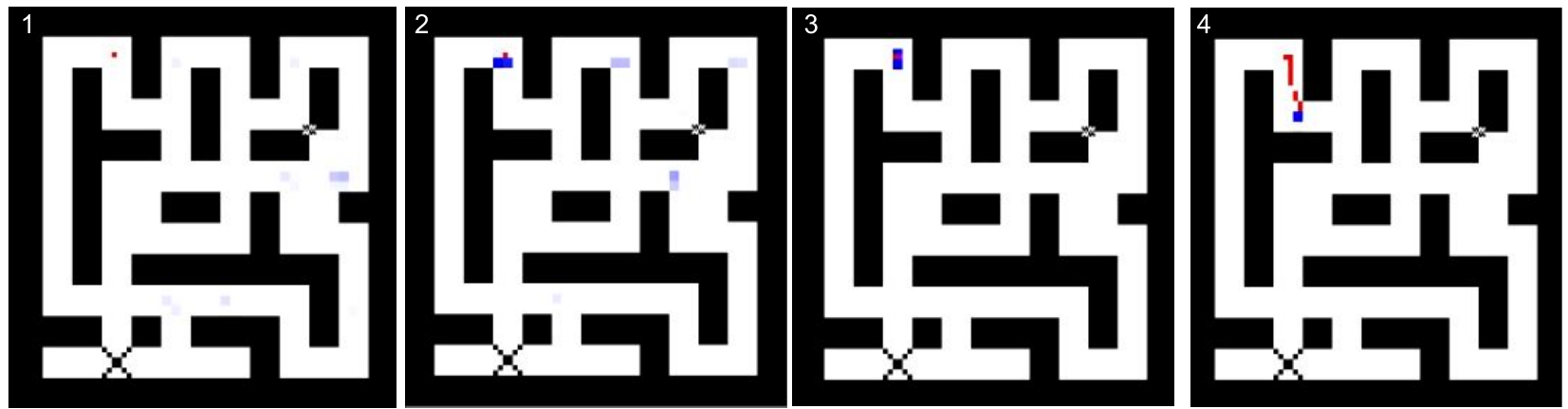}
\caption{Four example frames to illustrate the typical behavior of the agent: The red line is the trace of its actual position, while the shades of blue represent its position estimate. The darker the blue, the more confident the agent is to be in this location. Frame 1 shows the agent's true starting position as a red dot, frame 2 shows several similar locations identified after a bit of turning, in frame 3 the agent starts to understand the true location, and in frame 4 it has moved. 
}
\label{fig:localization}
\end{figure}

If the agent finds the exit it does so in almost shortest path manner, as can be seen in Figure~\ref{fig:trajectories}. However, the agent needs a considerable number of steps to localize itself. 
To evaluate this localization overhead, we trained an agent consisting solely of the reactive agent module with access to the perfect location and optimal short term target direction and plotted its average performance on the test set in Figure~\ref{fig:full_agent_test_comparison}. 
The figure shows a large gap between the full agent and the agent with access to the perfect position. This is due to turning actions, which the full agent performs to localize itself, i.e., it continuously needs to look around to know where it is.
For the localization in the beginning of an episode, the agent also mainly relies on turning as can be seen in four example frames in Figure~\ref{fig:localization}. 

\section{Conclusion}

We have presented a deep reinforcement learning agent that can localize itself on a 2D map based on observations of its 3D surroundings. The agent manages to find the exit in mazes with high success rate, even in mazes substantially larger than it has ever seen during training. The agent often finds the shortest path, showing that the agent can continuously retain a good localization.

The architecture of our system is built in a modular fashion. Each module deals with a subtask of the maze problem and is trained in isolation. This modularity allows for a structured architecture design, where a complex task is broken down into subtasks, and each subtask is then solved by a module. Modules consist of general architectures, e.g., MLPs, or more task-specific networks such as our recurrent localization cell. It is also possible to use deterministic algorithm modules, such as in our shortest path planning module. 
Architecture design is aided by the possibility to easily replace each module by ground truth values, if available, to find sources of bad performance. 

Our agent is designed for a specific task. We plan to make our modular architecture more general and apply it to other tasks, such as playing 3D games. Since modules can be swapped out and arranged differently, it would be interesting to equip an agent with many modules and let it learn which module to use in which situation.

\section*{Acknowledgments}
We would like to thank the anonymous reviewers for their helpful comments.

\appendix

\section{Visible Local Map Network Implementation Details}

The visible local map network passes the RGB visual input through two convolutional layers and a fully connected layer to extract visual features. These visual features are concatenated to the discretized angle input $\hat{\alpha}$ and passed through another fully connected layer to get an intermediate representation from which a local map excerpt and the currently visible field are estimated.  
The estimated map excerpt is constructed by passing the intermediate representation through a fully connected layer with a clipping nonlinearity that clips the output to $[-0.5,0.5]$. The visible field estimation is achieved as well by passing the intermediate representation through a fully connected layer, but here we use a rectified pseudo-sigmoidal nonlinearity such that each output component lies within $[0,1]$. We achieve this rectified pseudo-sigmoidal activation by clipping each output to $[-0.5,0.5]$ and then adding $0.5$. We use this rectified pseudo-sigmoidal activation because it is able to fully close (0) and open (1) the estimated gate. To get the visible local map estimation we multiply the estimated map excerpt component-wise with the visible field estimation, i.e., we gate the estimated map excerpt with the visible field estimation. Note that we do not directly train the network to estimate the visible field, but rather used this term to give a logical intuition why the gating operation is effective.

\section{Recurrent Localization Cell Implementation Details}

We estimate the egomotion $s_t$ as a discrete probability distribution over a 3x3 grid. The grid cells represent the estimated probability that the agent moved North-West, North, North-East and so on. The cell in the middle of the 3x3 grid represents the estimated probability that the agent stayed in place. We get a rough estimate of the egomotion by using the visible local map input as a two dimensional convolution filter and shift it over the previously estimated local map. This rough estimate is fine tuned by the outputs of a small feed forward neural network which takes as inputs the previously estimated egomotion probability distribution ($s_{t-1}$), the current discretized angle ($\hat{\alpha}$), the last action taken ($a_{t-1}$) and the last extrinsic reward received ($r_{t-1})$. We use a softmax over the estimated egomotion logits to get the estimated egomotion probability distribution $s_t$.

We then use the estimated egomotion probability distribution as two dimensional convolution filter to shift the previously estimated local map ($LM_{t-1}^{est}$) and a map feedback ($LM_{t-1}^{mfb}$) from the previously estimated position. We add the new visible local map to the shifted previously estimated map and clip it to the range $[-0.5,+0.5]$ to get the new local map estimation $LM_{t}^{est}$. The shifted map feedback is weighted by a trainable parameter $\lambda$ and added to the new local map estimation to get the local map estimation with map feedback $LM_t^{est+mfb}$, which is again clipped to $[-0.5,+0.5]$. Note that we differentiate between the estimated local map $LM_{t}^{est}$, which is recurrently passed to the next step, and the estimated local map with map feedback $LM_t^{est+mfb}$, which is only used for the immediate localization. We do not include the map feedback directly into the estimated local map since otherwise a map feedback from an incorrect position could alter the construction of the estimated local map. The map feedback is merely used to complete the estimated local map.

We rasterize and scale down the full map to fit the desired location granularity and range, i.e., we scale it to have $N$ discrete cells in the range $[-0.5,+0.5]$; $-0.5$ representing black, $+0.5$ representing white. To localize the agent, we then simply use the estimated local map with map feedback as two dimensional convolution filter and slide it over the zero-padded rasterized map to get for each possible location cell the correlation of the surrounding local map with the local map estimation. We use a softmax over the $N$ correlation outputs to get the location probability distribution $\{p_i^{loc}\}_{i=1}^N$. 

Finally, the feedback from the map is extracted for the next localization step: we get for each location cell in the rasterized map the corresponding local map surrounding it and sum up these $N$ local maps, each weighted by the probability of the agent being in the corresponding cell.

\section{Shortest Path Planning Algorithm}

We here present our implementation of the shortest path planning algorithm. Note that the deterministic shortest path planner was not the main focus of our work. Any algorithm that takes a grid maze as input and outputs for each location in the maze the shortest path direction for the next step to the nearest exit would work.

We first replaced the reward estimation in each location of the reward map with the average over the corresponding maze cell and used a sharp softmax to classify each location cell into either target cell, navigable cell or wall cell. We assigned the values $1.0$, $0.99$ and $0$ to cells that are target cells, navigable cells and wall cells respectively. We used this module in a recursive multiplicative way to assign each location cell a value $v$ corresponding to the distance to the nearest target cell. More precisely we did 200 iterations where in each iteration we calculated for each cell $i$:
$$v_k^i = v_{k-1}^i\cdot\max_{j\in\mathcal{N}_i}{v_{k-1}^j}$$

Here, $k\in\{1,...,200\}$ denotes the current iteration and $\mathcal{N}_i$ denotes the set of neighboring cells of $i$ to the North, East, South and West. In the final iteration, we return for each cell $i$ a sharp softmax over the four values in $\mathcal{N}_i$, which is the desired short term target direction and $1.0-v_k^i$, which is the desired measure of distance to the nearest target.

\begin{figure*}[htb]
\centering
\includegraphics[width=\textwidth]{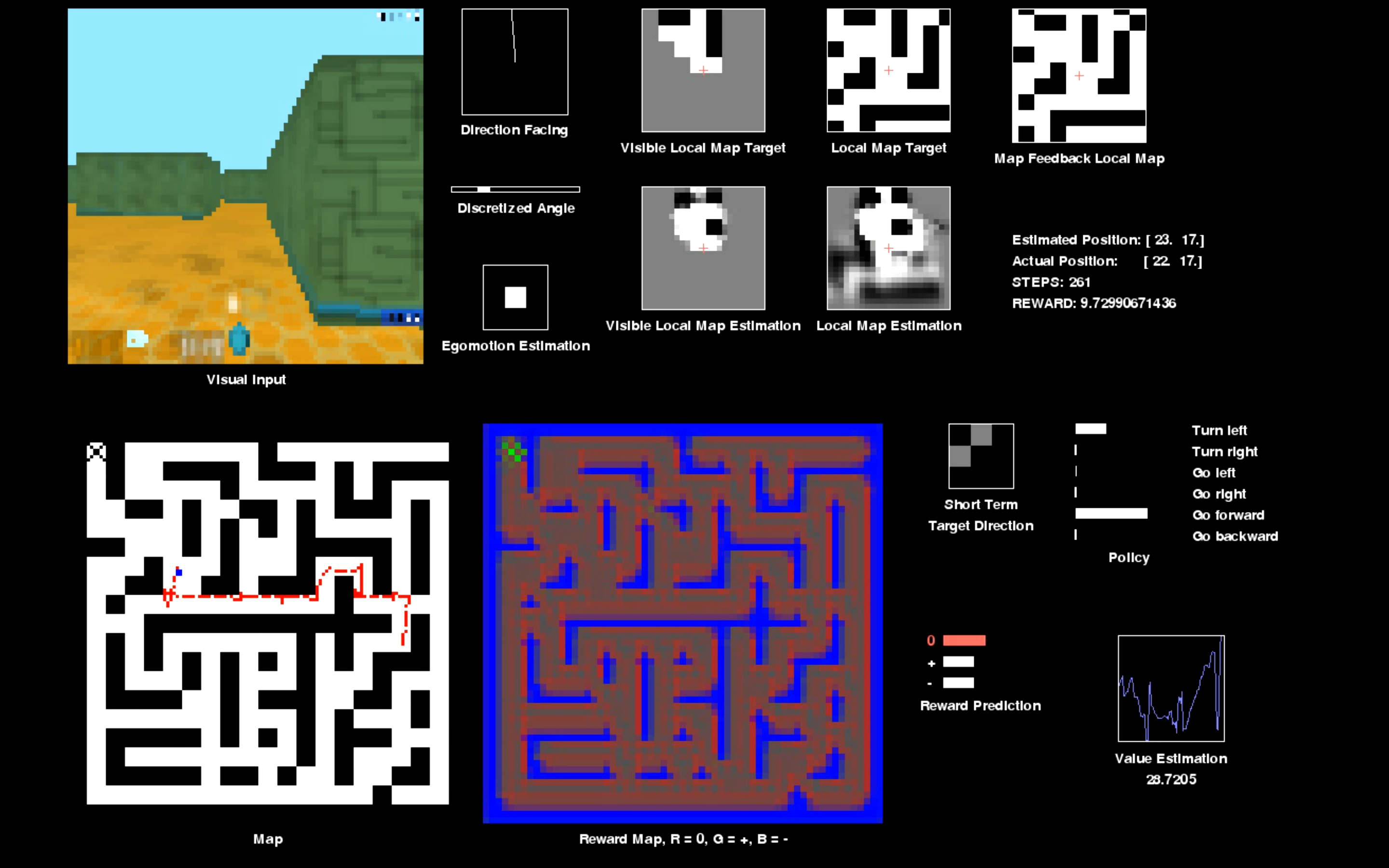}
\caption{Example of our agent moving through a maze of size 21x21 during testing. The \emph{Visual Input} is the 3D pixel input the agent receives. The \emph{Map} is the map of the maze showing the agent's position estimation (blue dot) and the walked trajectory (red trace). The agent only sees the \quotes{X}. The \emph{Reward Map} shows the agent's estimation of where to get reward. The displayed map is shows positive reward estimations in green, 0 reward estimations in red and negative reward estimations in blue. The agent has successfully learned that reaching the \quotes{X} gives positive reward, and walking into walls yields negative reward. On the top right we can see how the agent estimates its (visible) local map. It then uses the local map estimate to localize itself within the maze map. One can see that the agent is currently confident of his position (only a single blue dot in the map), and that its localization is indeed very accurate, as can be seen by comparing its estimated position ([23 17]) and its actual position ([22 17]). The bottom right shows the probability distribution over the possible actions for the current state (policy).}
\label{fig:fullEnvExample}
\end{figure*}

\section{Screenshot of Full Agent Navigating a Maze}

Figure \ref{fig:fullEnvExample} shows a screenshot of our agent in a maze from the test set.

\newpage
\newpage

\bibliography{mapreader}

\begin{thebibliography}{}

\bibitem[\protect\citeauthoryear{Barto, Sutton, and
  Anderson}{1983}]{DBLP:journals/tsmc/BartoSA83}
Barto, A.~G.; Sutton, R.~S.; and Anderson, C.~W.
\newblock 1983.
\newblock Neuronlike adaptive elements that can solve difficult learning
  control problems.
\newblock {\em {IEEE} Trans. Systems, Man, and Cybernetics} 13(5):834--846.

\bibitem[\protect\citeauthoryear{Beattie \bgroup et al\mbox.\egroup
  }{2016}]{DBLP:journals/corr/BeattieLTWWKLGV16}
Beattie, C.; Leibo, J.~Z.; Teplyashin, D.; Ward, T.; Wainwright, M.;
  K{\"{u}}ttler, H.; Lefrancq, A.; Green, S.; Vald{\'{e}}s, V.; Sadik, A.;
  Schrittwieser, J.; Anderson, K.; York, S.; Cant, M.; Cain, A.; Bolton, A.;
  Gaffney, S.; King, H.; Hassabis, D.; Legg, S.; and Petersen, S.
\newblock 2016.
\newblock Deepmind lab.
\newblock {\em CoRR} abs/1612.03801.

\bibitem[\protect\citeauthoryear{Bhatti \bgroup et al\mbox.\egroup
  }{2016}]{DBLP:journals/corr/BhattiDMNST16}
Bhatti, S.; Desmaison, A.; Miksik, O.; Nardelli, N.; Siddharth, N.; and Torr,
  P. H.~S.
\newblock 2016.
\newblock Playing doom with slam-augmented deep reinforcement learning.
\newblock {\em CoRR} abs/1612.00380.

\bibitem[\protect\citeauthoryear{Elfes}{1989}]{DBLP:journals/computer/Elfes89}
Elfes, A.
\newblock 1989.
\newblock Using occupancy grids for mobile robot perception and navigation.
\newblock {\em {IEEE} Computer} 22(6):46--57.

\bibitem[\protect\citeauthoryear{Fuentes{-}Pacheco, Ascencio, and
  Rend{\'{o}}n{-}Mancha}{2015}]{DBLP:journals/air/Fuentes-PachecoAR15}
Fuentes{-}Pacheco, J.; Ascencio, J.~R.; and Rend{\'{o}}n{-}Mancha, J.~M.
\newblock 2015.
\newblock Visual simultaneous localization and mapping: a survey.
\newblock {\em Artif. Intell. Rev.} 43(1):55--81.

\bibitem[\protect\citeauthoryear{Gullapalli}{1990}]{DBLP:journals/nn/Gullapalli90}
Gullapalli, V.
\newblock 1990.
\newblock A stochastic reinforcement learning algorithm for learning
  real-valued functions.
\newblock {\em Neural Networks} 3(6):671--692.

\bibitem[\protect\citeauthoryear{Gupta \bgroup et al\mbox.\egroup
  }{2017}]{DBLP:journals/corr/GuptaDLSM17}
Gupta, S.; Davidson, J.; Levine, S.; Sukthankar, R.; and Malik, J.
\newblock 2017.
\newblock Cognitive mapping and planning for visual navigation.
\newblock {\em CoRR} abs/1702.03920.

\bibitem[\protect\citeauthoryear{Jaderberg \bgroup et al\mbox.\egroup
  }{2016}]{DBLP:journals/corr/JaderbergMCSLSK16}
Jaderberg, M.; Mnih, V.; Czarnecki, W.~M.; Schaul, T.; Leibo, J.~Z.; Silver,
  D.; and Kavukcuoglu, K.
\newblock 2016.
\newblock Reinforcement learning with unsupervised auxiliary tasks.
\newblock {\em CoRR} abs/1611.05397.

\bibitem[\protect\citeauthoryear{Kaelbling, Littman, and
  Moore}{1996}]{DBLP:journals/jair/KaelblingLM96}
Kaelbling, L.~P.; Littman, M.~L.; and Moore, A.~W.
\newblock 1996.
\newblock Reinforcement learning: {A} survey.
\newblock {\em J. Artif. Intell. Res.} 4:237--285.

\bibitem[\protect\citeauthoryear{Kempka \bgroup et al\mbox.\egroup
  }{2016}]{DBLP:conf/cig/KempkaWRTJ16}
Kempka, M.; Wydmuch, M.; Runc, G.; Toczek, J.; and Jaskowski, W.
\newblock 2016.
\newblock Vizdoom: {A} doom-based {AI} research platform for visual
  reinforcement learning.
\newblock In {\em {IEEE} Conference on Computational Intelligence and Games,
  {CIG} 2016, Santorini, Greece, September 20-23, 2016},  1--8.

\bibitem[\protect\citeauthoryear{Kulkarni \bgroup et al\mbox.\egroup
  }{2016}]{DBLP:conf/nips/KulkarniNST16}
Kulkarni, T.~D.; Narasimhan, K.; Saeedi, A.; and Tenenbaum, J.
\newblock 2016.
\newblock Hierarchical deep reinforcement learning: Integrating temporal
  abstraction and intrinsic motivation.
\newblock In {\em Advances in Neural Information Processing Systems 29: Annual
  Conference on Neural Information Processing Systems 2016, December 5-10,
  2016, Barcelona, Spain},  3675--3683.

\bibitem[\protect\citeauthoryear{Minsky}{1954}]{minsky1954theory}
Minsky, M.~L.
\newblock 1954.
\newblock {\em Theory of neural-analog reinforcement systems and its
  application to the brain model problem}.
\newblock Princeton University.

\bibitem[\protect\citeauthoryear{Mirowski \bgroup et al\mbox.\egroup
  }{2016}]{DBLP:journals/corr/MirowskiPVSBBDG16}
Mirowski, P.; Pascanu, R.; Viola, F.; Soyer, H.; Ballard, A.~J.; Banino, A.;
  Denil, M.; Goroshin, R.; Sifre, L.; Kavukcuoglu, K.; Kumaran, D.; and
  Hadsell, R.
\newblock 2016.
\newblock Learning to navigate in complex environments.
\newblock {\em CoRR} abs/1611.03673.

\bibitem[\protect\citeauthoryear{Mnih \bgroup et al\mbox.\egroup
  }{2015}]{DBLP:journals/nature/MnihKSRVBGRFOPB15}
Mnih, V.; Kavukcuoglu, K.; Silver, D.; Rusu, A.~A.; Veness, J.; Bellemare,
  M.~G.; Graves, A.; Riedmiller, M.~A.; Fidjeland, A.; Ostrovski, G.; Petersen,
  S.; Beattie, C.; Sadik, A.; Antonoglou, I.; King, H.; Kumaran, D.; Wierstra,
  D.; Legg, S.; and Hassabis, D.
\newblock 2015.
\newblock Human-level control through deep reinforcement learning.
\newblock {\em Nature} 518(7540):529--533.

\bibitem[\protect\citeauthoryear{Mnih \bgroup et al\mbox.\egroup
  }{2016}]{DBLP:journals/corr/MnihBMGLHSK16}
Mnih, V.; Badia, A.~P.; Mirza, M.; Graves, A.; Lillicrap, T.~P.; Harley, T.;
  Silver, D.; and Kavukcuoglu, K.
\newblock 2016.
\newblock Asynchronous methods for deep reinforcement learning.
\newblock {\em CoRR} abs/1602.01783.

\bibitem[\protect\citeauthoryear{Peters and
  Schaal}{2008}]{DBLP:journals/nn/PetersS08}
Peters, J., and Schaal, S.
\newblock 2008.
\newblock Reinforcement learning of motor skills with policy gradients.
\newblock {\em Neural Networks} 21(4):682--697.

\bibitem[\protect\citeauthoryear{Rumelhart \bgroup et al\mbox.\egroup
  }{1988}]{rumelhart1988learning}
Rumelhart, D.~E.; Hinton, G.~E.; Williams, R.~J.; et~al.
\newblock 1988.
\newblock Learning representations by back-propagating errors.
\newblock {\em Cognitive modeling} 5(3):1.

\bibitem[\protect\citeauthoryear{Schaul \bgroup et al\mbox.\egroup
  }{2015}]{DBLP:journals/corr/SchaulQAS15}
Schaul, T.; Quan, J.; Antonoglou, I.; and Silver, D.
\newblock 2015.
\newblock Prioritized experience replay.
\newblock {\em CoRR} abs/1511.05952.

\bibitem[\protect\citeauthoryear{Sutton and
  Barto}{1998}]{DBLP:books/lib/SuttonB98}
Sutton, R.~S., and Barto, A.~G.
\newblock 1998.
\newblock {\em Reinforcement learning - an introduction}.
\newblock Adaptive computation and machine learning. {MIT} Press.

\bibitem[\protect\citeauthoryear{Sutton, Precup, and
  Singh}{1999}]{DBLP:journals/ai/SuttonPS99}
Sutton, R.~S.; Precup, D.; and Singh, S.~P.
\newblock 1999.
\newblock Between mdps and semi-mdps: {A} framework for temporal abstraction in
  reinforcement learning.
\newblock {\em Artif. Intell.} 112(1-2):181--211.

\bibitem[\protect\citeauthoryear{Sutton}{1984}]{sutton1984temporal}
Sutton, R.~S.
\newblock 1984.
\newblock Temporal credit assignment in reinforcement learning.

\bibitem[\protect\citeauthoryear{Sutton}{1988}]{DBLP:journals/ml/Sutton88}
Sutton, R.~S.
\newblock 1988.
\newblock Learning to predict by the methods of temporal differences.
\newblock {\em Machine Learning} 3:9--44.

\bibitem[\protect\citeauthoryear{Szepesv{\'{a}}ri}{2010}]{DBLP:series/synthesis/2010Szepesvari}
Szepesv{\'{a}}ri, C.
\newblock 2010.
\newblock {\em Algorithms for Reinforcement Learning}.
\newblock Synthesis Lectures on Artificial Intelligence and Machine Learning.
  Morgan {\&} Claypool Publishers.

\bibitem[\protect\citeauthoryear{Thrun, Burgard, and
  Fox}{2005}]{thrun2005probabilistic}
Thrun, S.; Burgard, W.; and Fox, D.
\newblock 2005.
\newblock {\em Probabilistic robotics}.
\newblock MIT press.

\bibitem[\protect\citeauthoryear{van Hasselt, Guez, and
  Silver}{2016}]{DBLP:conf/aaai/HasseltGS16}
van Hasselt, H.; Guez, A.; and Silver, D.
\newblock 2016.
\newblock Deep reinforcement learning with double q-learning.
\newblock In {\em Proceedings of the Thirtieth {AAAI} Conference on Artificial
  Intelligence, February 12-17, 2016, Phoenix, Arizona, {USA.}},  2094--2100.

\bibitem[\protect\citeauthoryear{Wang \bgroup et al\mbox.\egroup
  }{2016}]{DBLP:conf/icml/WangSHHLF16}
Wang, Z.; Schaul, T.; Hessel, M.; van Hasselt, H.; Lanctot, M.; and de~Freitas,
  N.
\newblock 2016.
\newblock Dueling network architectures for deep reinforcement learning.
\newblock In {\em Proceedings of the 33nd International Conference on Machine
  Learning, {ICML} 2016, New York City, NY, USA, June 19-24, 2016},
  1995--2003.

\bibitem[\protect\citeauthoryear{Watkins and Dayan}{1992}]{watkins1992q}
Watkins, C.~J., and Dayan, P.
\newblock 1992.
\newblock Q-learning.
\newblock {\em Machine learning} 8(3-4):279--292.

\bibitem[\protect\citeauthoryear{Watkins}{1989}]{watkins1989learning}
Watkins, C. J. C.~H.
\newblock 1989.
\newblock {\em Learning from delayed rewards}.
\newblock Ph.D. Dissertation, King's College, Cambridge.

\bibitem[\protect\citeauthoryear{Williams}{1992}]{DBLP:journals/ml/Williams92}
Williams, R.~J.
\newblock 1992.
\newblock Simple statistical gradient-following algorithms for connectionist
  reinforcement learning.
\newblock {\em Machine Learning} 8:229--256.

\bibitem[\protect\citeauthoryear{Zhu \bgroup et al\mbox.\egroup
  }{2017}]{DBLP:conf/icra/ZhuMKLGFF17}
Zhu, Y.; Mottaghi, R.; Kolve, E.; Lim, J.~J.; Gupta, A.; Fei{-}Fei, L.; and
  Farhadi, A.
\newblock 2017.
\newblock Target-driven visual navigation in indoor scenes using deep
  reinforcement learning.
\newblock In {\em 2017 {IEEE} International Conference on Robotics and
  Automation, {ICRA} 2017, Singapore, Singapore, May 29 - June 3, 2017},
  3357--3364.

\end{thebibliography}
\bibliographystyle{aaai}

\end{document}